\documentclass{article}

\usepackage{microtype}
\usepackage{graphicx}
\usepackage{booktabs} 
\usepackage{multirow}  
\usepackage{bigstrut} 
\usepackage[table]{xcolor}
\definecolor{lightgray}{gray}{0.9}
\usepackage{subcaption} 
\usepackage{enumitem} 
\usepackage{hyperref}

\usepackage[accepted]{icml2024}

\usepackage{amsmath}
\usepackage{amssymb}
\usepackage{mathtools}
\usepackage{amsthm}

\usepackage[capitalize,noabbrev]{cleveref}

\theoremstyle{plain}

\theoremstyle{definition}

\theoremstyle{remark}

\usepackage[textsize=tiny]{todonotes}

\newcommand{\ieno}{\textit{i}.\textit{e}.}
\newcommand{\egno}{\textit{e}.\textit{g}.} 

\definecolor{ForestGreen}{RGB}{0, 179, 45}
\newcommand{\tcb}{\textcolor{black}}

\begin{document}

\twocolumn[
\icmltitle{Slot-VLM: SlowFast Slots for Video-Language Modeling}



\icmlsetsymbol{equal}{*}


\begin{icmlauthorlist}
\icmlauthor{Jiaqi Xu}{equal,ustc}
\icmlauthor{Cuiling Lan}{msra}
\icmlauthor{Wenxuan Xie}{msra}
\icmlauthor{Xuejin Chen}{ustc}
\icmlauthor{Yan Lu}{msra} 
\end{icmlauthorlist}

\icmlaffiliation{ustc}{University of Science and Technology of China}
\icmlaffiliation{msra}{Microsoft Research Asia}


\icmlkeywords{Machine Learning, ICML}

\vskip 0.3in
]



\printAffiliationsAndNotice{\icmlEqualContribution} 

\begin{abstract}
Video-Language Models (VLMs), powered by the advancements in Large Language Models (LLMs), are charting new frontiers in video understanding. A pivotal challenge is the development of an efficient method to encapsulate video content into a set of representative tokens to align with LLMs. In this work, we introduce Slot-VLM, a novel framework designed to generate semantically decomposed video tokens, in terms of object-wise and event-wise visual representations, to facilitate LLM inference. Particularly, we design a SlowFast Slots module, \ieno, SF-Slots, that adaptively aggregates the dense video tokens from the CLIP vision encoder to a set of representative slots. In order to take into account both the spatial object details and the varied temporal dynamics, SF-Slots is built with a dual-branch structure. The Slow-Slots branch focuses on extracting object-centric slots from features at high spatial resolution but low (slow) frame sample rate, emphasizing detailed object information. Conversely, Fast-Slots branch is engineered to learn event-centric slots from high temporal sample rate but low spatial resolution features. These complementary slots are combined to form the vision context, serving as the input to the LLM for efficient question answering. Our experimental results demonstrate the effectiveness of our Slot-VLM, which achieves the state-of-the-art performance on video question-answering.

\end{abstract}

\section{Introduction}
\label{sec:introduction}
Recently, Large Language Models (LLMs) have gained significant progress \cite{brown2020language,touvron2023llama,openai2023chatgpt}. They present exceptional ability to comprehend, reason with, and generate human language text. Such amazing capabilities have stimulated the wave of research on extending the models to Vision Language Models, enabling the vision reasoning ability. 

\begin{figure}[t]
  \centering
  
   \includegraphics[width=1\linewidth]{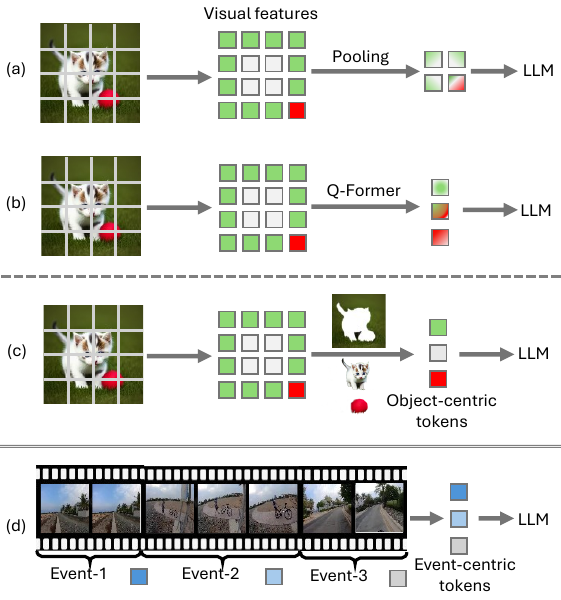}
   \vspace{-6mm}
   \caption{Illustration of methods for aligning visual features with LLM. Previous methods (a) and (b) leverage pooling or Q-Former to aggregate visual tokens, where each generated token contains \emph{coupled} semantics. In contrast, we propose to generate semantically decoupled object-centric tokens as illustrated in (c), and event-centric tokens as illustrated in (d), to align with the LLM. } 
   \vspace{-2mm}
   \label{fig:idea}
\end{figure}

\begin{figure*}[t]
  \centering
   \includegraphics[width=1\linewidth]{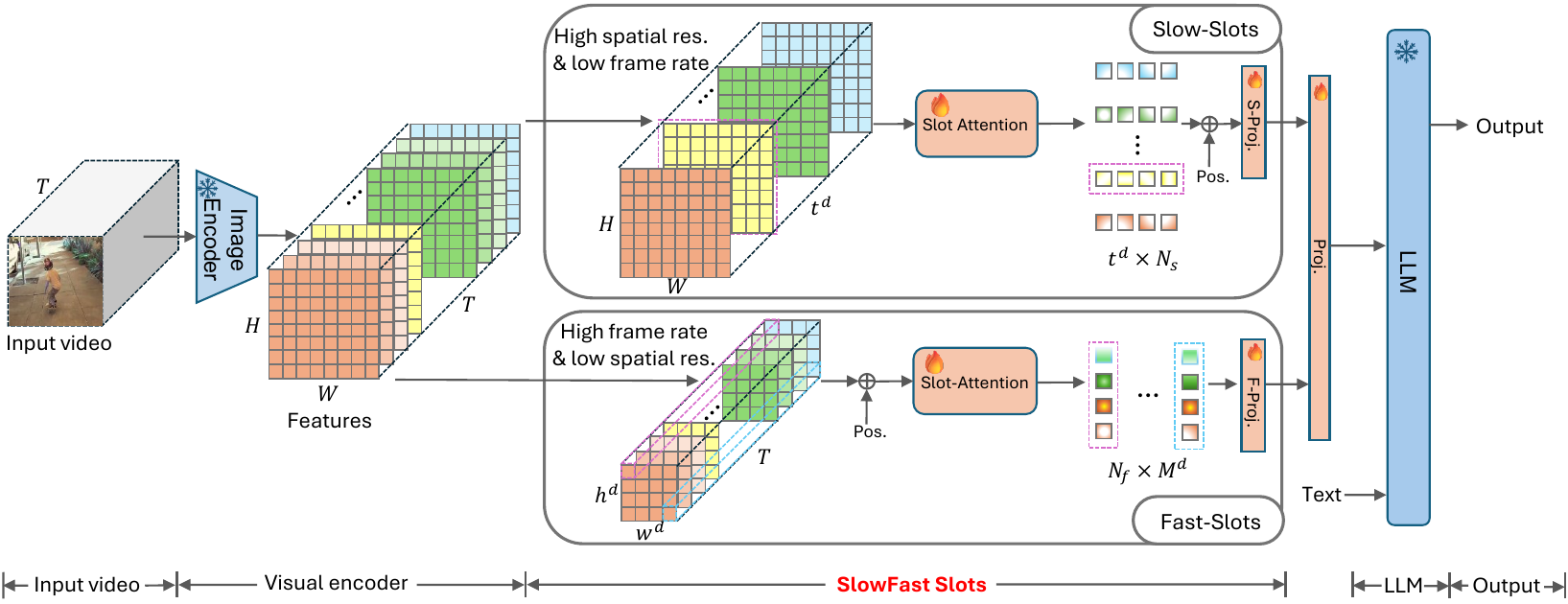}
   \caption{Flowchart of our proposed Slot-VLM for video understanding. Slot-VLM consists of a frozen image encoder, a learnable SlowFast Slots module (\ieno, SF-Slots module), a projection layer, and a frozen LLM. The image encoder (CLIP image encoder) encodes the input video of $T$ frames into a sequence of image features, resulting in extensive ($H\times W \times T$) video tokens. In order to obtain semantically decoupled and compact (reduced) video tokens as the vision context for aligning with LLM, our SlowFast Slots module learns to aggregate those tokens to object-centric tokens and event-centric tokens through the Slow-Slots branch and the Fast-Slots branch, respectively. The Slow-Slots branch operates at low frame rate ($t^d \ll T$) but high spatial resolution in order to capture spatial objects through slot attention on each frame. The Fast-Slot branch operates at high frame rate but low spatial resolution ($M^d= h^d \times w^d, h^d < H$, $w^d < W$) in order to capture temporal dynamics through slot attention over each spatial position. The learned slots (tokens) from two branches are projected by a fully connected layer and input to LLM for video reasoning, together with the text input (text query). } 
   \label{fig:framework}
\end{figure*}

For image understanding, MiniGPT-4 \cite{zhu2023minigpt} leverages a Q-Former and a projector to align a frozen visual encoder with a frozen advanced LLM, where Q-Former converts the visual input into a fixed-length learned visual query tokens. LLaVA \cite{liu2023visual} and MiniGPT-v2 \cite{chen2023minigpt} directly use the gridded visual tokens (after projection) as the LLM input. Image-text pairs are leveraged to align the visual model and the LLM in training. For handling videos, one straightforward way is to stack the tokens from sampled frames and feed them into the LLM \cite{li2023blip, dai2305instructblip,zhao2023chatbridge}. This is challenging when we densely sample frames (\egno, with the purpose of preserving details) or when we sample abundant frames for long videos. For example, for a video of 10 minutes, when we sample at 1 frame per second and each frame uses 32 tokens, we will have 19,200 tokens in total. On the one hand, the large number of tokens increases both the memory and computational requirement. On the other hand, there is spatio-temporal redundancy and the features are not representative. 

Video-ChatGPT~\cite{maaz2023video}  reduces the number of video tokens to 356 by separately performing spatial pooling and temporal pooling on the video features, which suffers from the loss of vision details. Video-LLaMA \cite{damonlpsg2023videollama} and VideoChat \cite{li2023videochat} compress the video tokens using Q-Former, where a set of learnable queries aggregate the information from video tokens through cross-attention and self-attention. These strategies generate reduced tokens. However, each token contains semantically mixed/entangled information\footnote{See the visualization analysis of Q-Former in Appendix \ref{subsec:vis-QFormer}.} and this may bring inefficiency to video-language modeling. As we know, the text words as the input to LLMs have semantics and are disentangled. Intuitively, to better align the video tokens with the language input of LLMs for building VLMs, the synthesized video tokens should act similarly to words.  

In this work, as illustrated in Figure~\ref{fig:idea}, we aim to generate semantic-centric tokens\footnote{
A semantic-centric token refers to a token that represents a semantically meaningful entity, such as an object, or an event (from a temporally consistent segment).} from video features to comply with LLMs for efficient video-language modeling. By leveraging slot attention \cite{locatello2020object,seitzer2022bridging}, which converts an image into object-centric representations, \ieno, slots, we design a SlowFast Slots module (\ieno, SF-Slots module), that generates a set of semantically decomposed video tokens (\emph{a.k.a.} slots) from video features and takes such slots as input to LLMs. This is somewhat similar to human visual reasoning, where the brain converts the visual perception to high-level object representations to facilitate further reasoning. We dub the scheme powered by SF-Slots module as Slot-VLM. Figure~\ref{fig:framework} shows the flowchat of Slot-VLM. Particularly, SF-Slots module consists of two branches: Slow-Slots and Fast-Slot for decomposed spatial and temporal modeling, focusing on spatial objects and temporal varied dynamics (event), respectively. The Slow-Slots branch extracts object-centric slots from high spatial resolution features but sampled at low frame rate. The Fast-Slot branch extracts event-centric slots from high temporal resolution but low spatial resolution features. The two sets of slots are combined together as the input to the LLM for video reasoning. The visual encoder and the LLM are frozen. In instruction-tuning, we fine-tune SF-Slots module and the projection layer to align visual features with LLM.

In summary, we have three main contributions:
\begin{itemize} [noitemsep,nolistsep,leftmargin=*]
\vspace{-2mm}
    \item We propose a new framework, Slot-VLM, for efficient video-language modeling, where we generate semantically decomposed tokens to comply with LLMs. To our best knowledge, this is the first work that explores using learnable semantically decoupled tokens for aligning with LLM.
    \item We design a SF-Slots module that encourages object-centric and event-centric visual represenations for modeling spatial objects and temporal dynamics. This moves a step towards the compact and semantically meaningful token representations in VLMs.
    \item Experimental results demonstrate the effectiveness of our SF-Slots designs, achieving state-of-the-art performance on video question answering tasks. Thanks to the decoupling-aware
    designs, our token generation is more efficient than Q-Former-based
    designs.
\end{itemize}
We \tcb{hope that} this work will inspire more designs towards generating efficient and semantically interpretable tokens for video-language modeling.   





\section{Related Work}
\label{sec:related}
\noindent\textbf{Visual Language Models} In considering the strong comprehension, reasoning, and generation ability of the LLMs, there is a surge of investigations on exploiting LLMs and vision encoders to build
visual-language models, enabling the vision reasoning ability \cite{alayrac2022flamingo,li2023blip,zhu2023minigpt,liu2023visual,li2023videochat,maaz2023video,zhang2023video,damonlpsg2023videollama, song2023moviechat}.

BLIP-2 \cite{li2023blip} leverages frozen pre-trained CLIP image encoders \cite{radford2021learning} and frozen LLMs to bootstrap language-image pre-training, with the modality gap bridged by a Querying Transformer (Q-Former) and a projector. Q-Former extracts a fixed number (\egno, 32) of output features (tokens) from the image encoder, acting as an information bottleneck to feed most useful information to the LLM. Other works like MiniGPT-4~\cite{zhu2023minigpt} and Video-LLaMA~\cite{damonlpsg2023videollama} similarly leverage Q-Former to encode visual tokens, followed by image-text or video-text pair instruction tuning. 

Compared with image reasoning, it is more challenging for video-language modeling, where the excessive number of visual tokens from video would enlarge the computation and memory cost to the LLM inference and thus constrains the application of VLMs. Some works sparsely sample frames and concatenate the tokens from these frames as the input to LLMs \cite{li2023blip, dai2305instructblip,zhao2023chatbridge}. However, this brings difficulty for understanding long videos or videos with high temporal dynamics, with high risk of losing temporal information. 

Video-ChatGPT \cite{maaz2023video} reduces the number of visual tokens by performing spatial and temporal pooling separately. LLaMA-VID \cite{li2023llama} represents each frame with two distinct tokens: context token and content token, where context token is obtained by aggregating the visual features based on the text query while the content token is obtained by pooling the features. 
Being task-agnostic, pooling would result in loss of helpful information. 
Moreover, the semantic-agnostic pooling inevitably results in the mixing of features of different semantics, where such coupled representations may bring difficulty for video reasoning. 
Q-Former provides an elegant way for generating reduced number of tokens. However, we found the learned tokens are still coupled, with abundant overlapped information among them. When aligning with the ``disentangled" word text tokens, such entanglement may result in poor compatibility with LLMs. \tcb{Chat-UniVi \cite{jin2023chat} merges the dense visual tokens by clustering and uses the averaged token features in a cluster as the cluster representation. With clustering being parameter-free and task-agnostic, this is sub-optimal.} It is still an open question on how to generate video tokens for constructing efficient VLMs. 

In this work, we explore the use of semantically decoupled tokens by learning spatial object-centric slots and temporal event-centric slots for video representation. We reveal that using semantically decoupled tokens as the input to LLMs are  effective for video-language modeling.


\noindent\textbf{Object-Centric Learning} Humans naturally decompose the observed environment into entities at the appropriate level of abstraction to act in the world \cite{seitzer2022bridging}. To obtain semantically decoupled representations, object-centric representation learning has attracted much interests \cite{burgess2019monet, locatello2020object, singh2021illiterate, seitzer2022bridging}. Slot Attention \cite{locatello2020object} is a representative work. A slot attention module is designed to interface with features to produces a set of abstract representations, named as slots, where each slot representing an object or a scene (background).
It demonstrates encouraging object decomposition ability in simulated data but fails on real-world images \cite{locatello2020object}.  
To successfully learn object-centric slots in real-world images, DINOSAUR \cite{seitzer2022bridging} learns the slots from pre-trained features by reconstructing the features. Such pre-trained features can be obtained from self-supervised learning techniques like DINO \cite{caron2021emerging}, MAE \cite{he2022masked}, or vision language pre-training like CLIP \cite{radford2021learning}. These features usually have a high level of homogeneity within objects and thus facilitate the forming of object-centric representations.     

In this work, we leverage slot attention to learn semantically decoupled tokens as the input to LLMs and investigate the effectiveness of aligning ``concept tokens" with LLM input.
Note that, interestingly, words in human language are abstractions of concepts, with different words tending to be disentangled.
We intend to bridge the connection between vision and language by representing video with ``concepts" in terms of object-centric slots and event-centric slots. 

\section{Proposed Slot-VLM}
\label{sec:method}

\subsection{Overview}
As illustrated in Figure~\ref{fig:framework}, our Slot-VLM consists of a frozen image encoder, a trainable SF-Slots module that contains two branches, a projection layer, and a frozen LLM. 
Given a video sequence, we extract frames at a speed of 1 frame per second (FPS) and obtain a video $V$ of $T$ frames. The video $V$ together with the text question (user instruction) is input to Slot-VLM, and Slot-VLM outputs the generated text answer. 

In order to build an effective VLM for video reasoning, we propose to encapsulate the dense video tokens into a small set of semantically decoupled tokens to align with the LLM. Our SF-Slots module  converts the dense video tokens into a set of object-centric slots (tokens) through the Slow-Slots branch, and a set of event-centric slots (tokens) through the Fast-Slot branch, which are then projected as the input to LLM for video reasoning.

Interestingly, our exploitation of object-centric vision representation for LLM reasoning shares some similar logic with how humans perform visual reasoning. Human visual reasoning is a complex cognitive process. Visual signals are processed initially through the primary visual cortex (V1) and subsequently integrated in higher-level visual processing areas, resulting in the formation of complex visual object representations. Such high-level object representations together with brain-stored knowledge are then used for logical reasoning and inference to interpret observations \cite{riesenhuber1999hierarchical,bar2004visual}. Similarly, Slot-VLM generates object-centric and event-centric token representations to provide the vision source for the LLM reasoning, where the LLM stores rich knowledge and has strong reasoning ability, and interpretable-text generation ability. Such exploration makes us move a small step towards human-brain like visual reasoning. 




\subsection{Visual Feature Extraction}

To extract visual features of a video, we simply use an image-based model to get per-frame feature. Following \citet{maaz2023video}, we utilize vision-language pre-trained model CLIP (ViT-L/14) \cite{radford2021learning} as our image encoder. For each frame, the image encoder outputs $H\times W$ visual tokens, with the dimension of a token denoted by $D$, where $H=W=16$ for the CLIP encoder. For a sampled video of $T$ frames, we have $H\times W \times T$ visual tokens. Taking a video of 3 minutes as an example, the total number of video tokens is $16\times 16 \times 180 = 46,080$. Such huge number of tokens would result in high computational and memory cost to the LLM inference. Moreover, this also exceeds the capacity of existing LLMs. Considering the spatial structure and and temporal dynamics, there are still large redundancy from the dense tokens. 

\subsection{SlowFast Slots (SF-Slots) Module}

Previous methods that aggregate the video tokens by pooling \cite{maaz2023video} or Q-Former \cite{zhao2023chatbridge, zhang2023video} actually generate tokens with semantics still entangled in each token, without targeting at semantics decoupled representations. In contrast, the word tokens, which are the input to LLMs, are more semantically decoupled. In this work, we explore the using of semantics-decoupled video representations for video-language modeling.

We design a SF-Slots module that encapsulates the dense video tokens from the visual encoder into a small set of semantic-centric tokens to facilitate the LLM inference. It is challenging to learn semantics-centric representations from the dense video features. DINOSAUR \cite{seitzer2022bridging} uses slot attention on the pre-trained image feature to learn object-centric representations without supervision. It is challenging to directly extend the slot learning onto video features, where the large number of video tokens (\egno, 46,080 for a video of 3 minutes) raises high memory and computation requirements. 

The spatiotemporal orientations of a video are not equally likely
\cite{feichtenhofer2019slowfast}. To efficiently capture the informative information for action recognition, \citet{feichtenhofer2019slowfast} introduce SlowFast network, which uses a Slow pathway to operate at low frame rate to capture spatial semantics, and a Fast pathway to operate at high frame rate to capture motion. Their Fast pathway is made very lightweight by reducing its channel capacity. In our work, in order to efficiently learn semantics-centric representations, we design SlowFast Slots module, where a Slow-Slots branch operates at low frame rate to capture spatial object-wise slots, and a Fast-Slots branch operates at high frame rate to capture temporal event-wise slots. 

As illustrated in Figure~\ref{fig:framework}, for the Slow-Slots branch, we perform slot attention learning on each sampled frame to obtain spatial slots (object-centric slots), and concatenate the slots of the sampled frames as the Slow-Slots branch slots. For the Fast-Slots branch, for each spatial position, we perform slot attention learning along the time to obtain temporal slots (event-centric slots). The temporal slots for all the spatial positions are concatenated as the Fast-Slots branch slots. Note that we perform temporal slot learning on the high frame rate but low spatial resolution features in order to reduce the number of  slots, which is proportional to the spatial resolution.

\noindent\textbf{Slow-Slots Branch} For a video of $H\times W \times T$ dense video tokens, we sample the features at low frame rate but high spatial resolution to obtain $H\times W \times t^d$ video tokens\footnote{Superscript $d$ indicates down-sampled.} . We set $t^d = 8$, \ieno, we uniformly sample 8 frames. For the $i^{th}$ frame, we have a set $\mathcal{S}_i$ of $M_s = H\times W$ features (tokens). $\mathcal{S}_i$ is taken as the input to slot attention module \cite{locatello2020object,seitzer2022bridging} to generate $N_s$ object-centric slots $\mathcal{O}_i = \{\mathbf{o}_{i,1}, \ldots, \mathbf{o}_{i,N_s}\}$ (we also refer to them as spatial slots). 

Slot attention uses an iterative mechanism to map from the input tokens to the slots. At each iteration, slots attention uses cross attention with attention coefficients that are normalized over the slots (where slots are the queries) to aggregate token information. The normalization over the slot introduces competition between the slots for promoting the forming of decoupled representations. 

To distinguish different frames, we add learnable temporal position embedding $\mathbf{p}_i$ to each slot of the $i^{th}$ frame and obtain the updated object slots as $\mathcal{O}_i = \{\mathbf{o}_{i,1}, \ldots, \mathbf{o}_{i,N_s}\}$, where $\mathbf{o}_{i,j} = \mathbf{o}_{i,j} + \mathbf{p}_i$. We concatenate the slots of all the $t^d$ frames and obtain $t^d \times N_s$ slots $\mathcal{O} = \{O_1, \ldots, O_{t^d}\} = \{\mathbf{o}_{1,1}, \mathbf{o}_{1,2}, \ldots, \mathbf{o}_{1,N_s}, \ldots, \mathbf{o}_{t^d,N_s}\}$. A linear projection layer (S-Proj.) transforms each token to facilitate the alignment with slots from the Fast-Slots branch.

\noindent\textbf{Fast-Slots Branch} For a video of $H\times W \times T$ dense video tokens, we sample the features at high frame rate but low spatial resolution to obtain $h^d\times w^{d} \times T$ video tokens. We obtain the spatial down-sampled tokens by averaging pooling with a stride of 4, therefore $h^d = H/4 = 4$ and $w^d = W/4 = 4$. 

As illustrated in Figure~\ref{fig:framework}, we perform slot learning along temporal tokens. To be aware of different frames, we add learnable temporal position embedding to each token. For the $k^{th}$ spatial position, where $k=1, \ldots, M^d$ and $M^d = h^d\times w^{d}$, we have a set $\mathcal{F}_k$ of $T$ features (tokens) from $T$ frames. $\mathcal{F}_k$ is taken as the input to slot attention module to generate $N_f$ event-centric slots $\mathcal{E}_k = \{\mathbf{e}_{k,1}, \ldots, \mathbf{e}_{k,N_f}\}$ (we also refer to them as temporal slots).  Note that for a given spatial position from $h^d\times w^{d}$ features, this corresponds to a large local patch in the pixel space, \ieno, 56$\times$56 patch for a video of 224$\times$224 spatial resolution. When observing the temporal dynamics of such a spatial position, the changes of the dynamics can tell evolution of events. Temporal slots aggregates the temporal tokens for explaining parts of the tokens, similar to identifying events. Thus, we name the learned slots as event-centric slots.   

We concatenate the slots from each of the $M^d$ spatial positions and obtain $M^d \times N_f$ slots $\mathcal{E} = \{\mathcal{E}_1, \ldots, \mathcal{E}_{M^d}\} = \{\mathbf{e}_{1,1}, \mathbf{e}_{1,2}, \ldots, \mathbf{e}_{1,N_f}, \ldots, \mathbf{e}_{M^d,N_f}\}$. A linear projection layer (F-Proj.) transforms each token to facilitate the alignment with slots from the Slow-Slots branch. 

\subsection{Connection to LLM}

We concatenate the object-centric slots from the Slow-Slots branch and the event-centric slots from the Fast-Slots branch and obtain $N= t^d \times N_s + M^d \times N_f $ slots. A linear projection layer (Proj.) transforms each token to align with the LLM.  The projected tokens and the text instruction (question) are taken as the input to LLM for video reasoning. We use the fine-tuned Vicuna-7B from LLaVA as our LLM.

\subsection{Training Strategy}
We divide the training procedure into three stages: slot attention pre-training, single branch instruction tuning, and two branch joint instruction tuning.

We first pre-train the \tcb{two} slot attention modules \tcb{in Slow-Slots branch and in Fast-Slots branch} separately,
by reconstructing their input features, respectively. This promotes the formation of object-centric and event-centric slot representations. Similar to DINOSAUR \cite{seitzer2022bridging}, we use transformer decoder to reconstruct features.

Then, we perform instruction tuning on the single branch schemes, separately. For example, we fine-tune the Slow-Slot branch and the projection layers using the instruction pairs from \citet{maaz2023video}, by loading the parameters from the first stage.

Finally, we fine-tune the SF-Slots module and the projection layer using the instruction pairs from \citet{maaz2023video}, by loading the parameters from SF-Slots module from the second stage.


\begin{table*}[htbp] 
  \centering
  \caption{Comparison with the state-of-the-art methods for video QA. All these models use Vicuna-7B as the LLM. Different methods may use different datasets for pre-training. Moreover, for the instruction tuning, different methods adopt different instruction data as illustrated in the second column. For example, 11K(V)+5.1M(I) denotes the instruction data comprises about 11,000 pairs of video instructions pairs and 5.1 million pairs of image instructions. Connector denotes the method for connecting the vision features and the LLM.}
  \resizebox{\textwidth}{!}{ 
    \begin{tabular}{c|c|c|cc|cc|cc}
    \hline
    \multirow{2}{*}{Model} & \multirow{2}{*}{\parbox{3cm}{\centering Instruction Data \\ (\# of Pairs)}} & \multirow{2}{*}{Connector} & \multicolumn{2}{c|}{MSVD-QA} & \multicolumn{2}{c|}{MSRVTT-QA} & \multicolumn{2}{c}{ActivityNet-QA} \bigstrut[t]\\
          &       &       & Acc.  & Score & Acc.  & Score & Acc.  & Score \bigstrut[b]\\
    \hline
    Video LLaMA \cite{damonlpsg2023videollama} & 11K(V)+5.1M(I) & Q-Former & 51.6  & 2.5   & 29.6  & 1.8   & 12.1  & 1.1 \bigstrut[t]\\
    Video Chat \cite{li2023videochat} & 11K(V)+7K(I) & Q-Former & 56.3  & 2.8   & 45    & 2.5   & 26.5  & 2.2 \\
    \rowcolor{lightgray} Video-ChatGPT \cite{maaz2023video} & 100K(V) & Pooling & 64.9  & 3.3   & 49.3  & 2.8   & 35.2  & 2.7 \\
    Chat-UniVi \cite{jin2023chat} & 2M(V)+433K(I) & Clustering & 65    & 3.6   & 54.6  & 3.1   & 45.8  & 3.2 \\
    \rowcolor{lightgray} Video-LLaVA \cite{lin2023video} & 100K(V) & -    & 64.8  & -     & 58.3  & -     & 40.7  & - \\
     Video-LLaVA \cite{lin2023video} & 100K(V)+665K(I) & -    & 70.7  & \textbf{3.9}   & \underline{59.2}  & \textbf{3.5}   & 45.3  & 3.3 \\
    \rowcolor{lightgray} BT-Adapter \cite{liu2023one} & 100K(V) & Temporal Adaptor & 67    & 3.6   & 51.2  & 2.9   & 46.1  & 3.2 \\
    LLaMA-VID \cite{li2023llama} & 100K(V)+625K(I) & Q-Former\&Pooling & 69.7  & 3.7   & 57.7  & 3.2   & 47.4  & 3.3 \\
    VideoChat2 \cite{li2023mvbench} &  0.8M(V)+1.1M(I) & Q-Former & 70    & \textbf{3.9} & 54.1  & \underline{3.3}   & \textbf{49.1} & 3.3 \bigstrut[t]\\
    MovieChat \cite{song2023moviechat} & 11K(V)+5.1M (I) & Merge+Q-Former & \textbf{75.2} & 3.8   & 52.7  & 2.6   & 45.7  & 3.1 \bigstrut[b]\\
    \hline
    \rowcolor{lightgray}Slot-VLM (Ours) & 100K(V) & SlowFast Slots & \underline{74.9}  & \underline{3.8}   & \textbf{69.7} & \underline{3.4} & \underline{48.3}  & \textbf{3.4} \bigstrut\\
    \hline
    \end{tabular}%
    }
  \label{tab:SOTA}%
\end{table*}%

\begin{figure*}[t]
  \centering
   \includegraphics[width=1\linewidth]{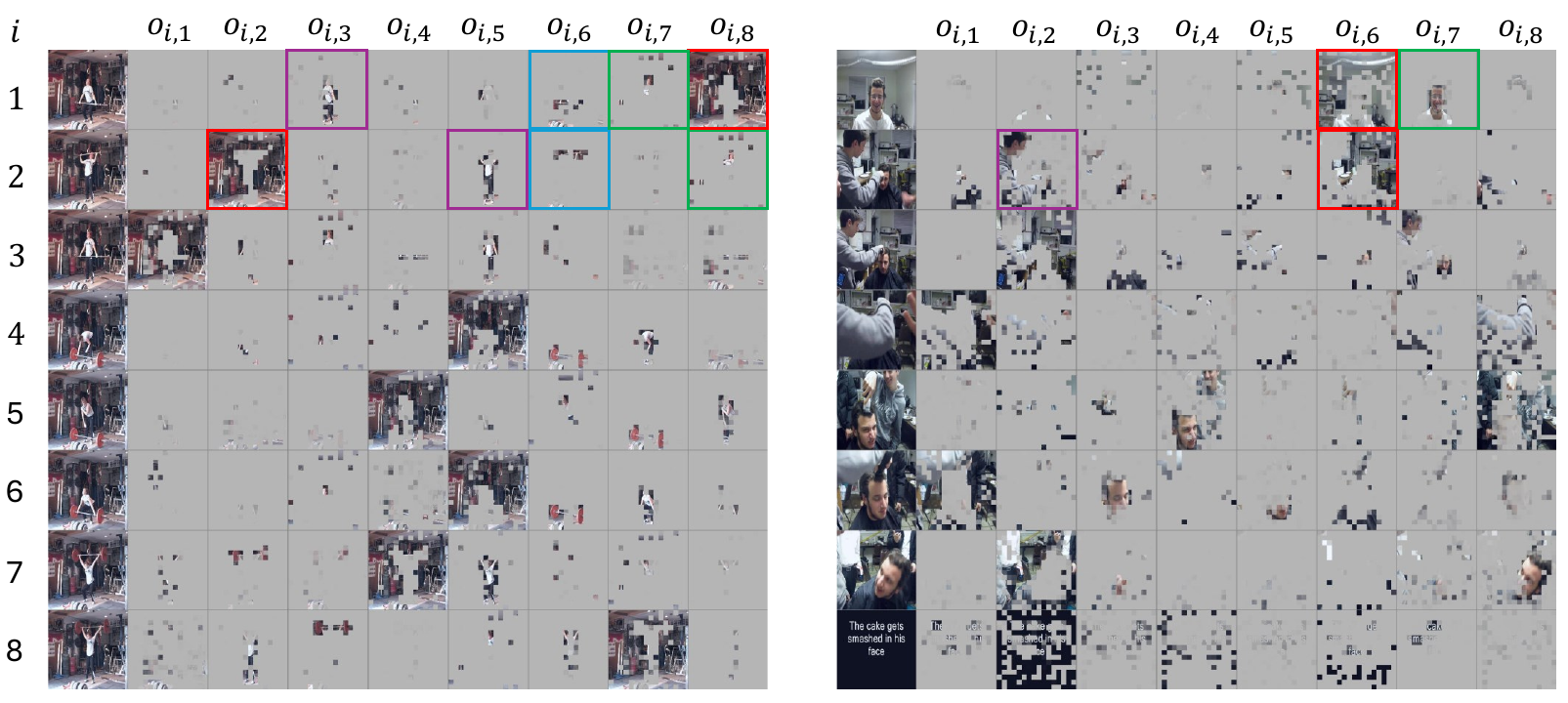}
   \caption{Visualization of spatial attention masks from the Slow-Slots branch for two video examples, respectively. We have $t^d=8$ frames as shown in 8 rows, indexed by $i$, where $i=1,\ldots,t^d$, respectively. The first column shows the original frame. The second to the ninth columns show the cross attention mask (from slot attention) for the $N_s=8$ object-centric slots $\mathcal{O}_i = \{\mathbf{o}_{i,1}, \ldots, \mathbf{o}_{i,N_s}\}$. We can see that even though not perfectly segmented, some meaningful slots have been formed. For example, the slots marked by red, purple, green, and blue in the first video (left) correspond to ``background", ``human body", ``head", and ``barbell". Note that the slots in a frame is unordered and exchangeable. } 
   \label{fig:visualization-S-mark}
\end{figure*}

\begin{figure*}[t]
  \centering
   \includegraphics[width=1\linewidth]{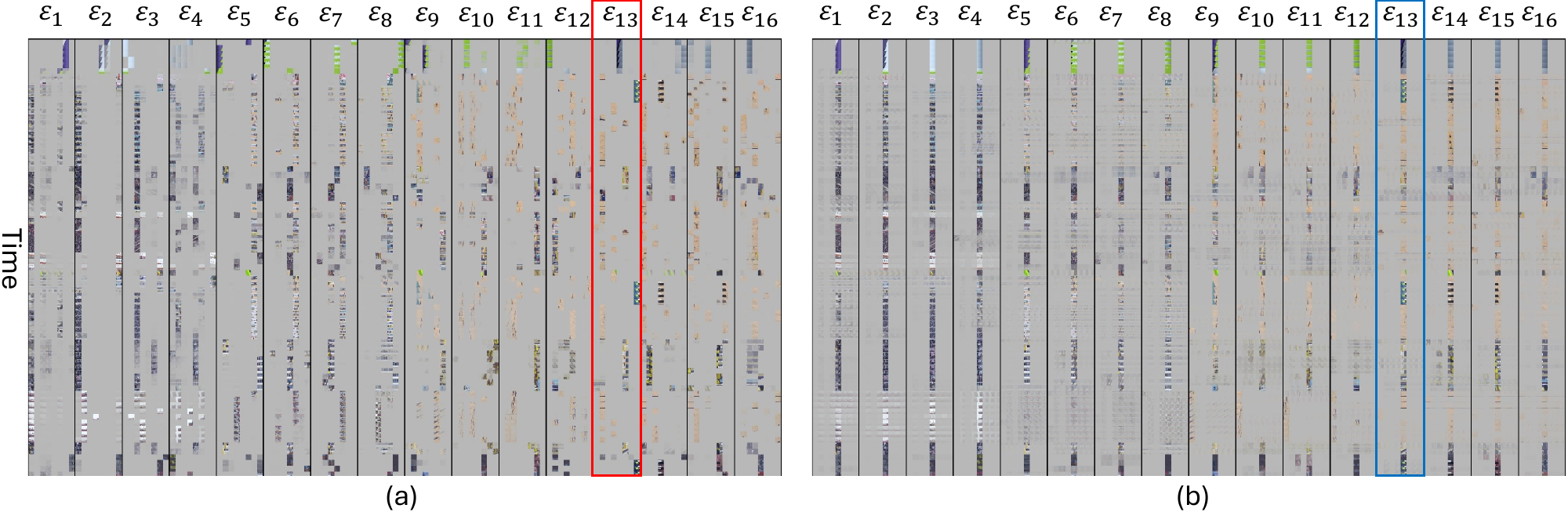}
   \caption{Visualization of temporal attention mask for $M^d = h^d \times w^d = 16$ spatial positions from (a) our Fast-Slots branch and (b) Fast-QFormer-VLM, respectively. \tcb{For simplicity, we also refer to slot as query here.} For the $k^{th}$ spatial position, we denote the set of learned temporal queries by $\mathcal{E}_k$. Take the $13^{th}$ spatial position of the query set $\mathcal{E}_{13}$ as an example (as marked by red box in (a) and blue box in (b)). For this spatial position, the models generate $N_f=8$ slots/queries by aggregating the temporal visual tokens.
   \tcb{The attention masks for $\mathcal{E}_{13}$ are denoted by a map of $T$ rows and $N_f$ columns,}
   with the visibility indicating which queries this
   temporal
   position belongs to.
   The higher the visibility, the greater the affinity between this temporal position and the query.
   We can see that in our Slot-VLM, similar contents
   tend
   to be allocated to the same slot, \ieno, different slots capture different contents (events) and present decoupled semantics. In contrast, in Fast-QFormer-VLM, different contents are usually assigned to the same query or are uniformly assigned to different queries. Note that for Fast-QFormer-VLM, we only show the mask of one head to save space, where similar observations can be found from other heads. A glimpse of the original video can be found in Appendix \ref{subsec:video-glimpse-temp}. See Figure~\ref{fig:Visualization-T-Sep-Enlarged} in Appendix \ref{subsec:video-glimpse-temp} for the enlarged visualization of $\mathcal{E}_{13}$.  } 
   \label{fig:Visualization-T-Sep}
\end{figure*}

\section{Experiments}
\label{sec:experiment}

\subsection{Implementation Details}
Following Video-ChatGPT \citep{maaz2023video}, for visual encoding, we employ the pre-trained CLIP ViT-L/14 model \citep{radford2021learning}, from which we derive our visual tokens. Note that the model size of CLIP ViT-L/14 (which we use) is 303M which is much smaller than CLIP ViT-G/14 (1012M). Specifically, we extract features from the penultimate layer, yielding an array of visual tokens with dimensions $H \times W$ for each video frame. We use the Vicuna (7B) from the LLaVA model \citep{liu2023visual}, to serve as our LLM. Both the image encoder and LLM are frozen in our training.
We sample at 1 fps from a video to obtain $T$ frames, and we resize each frame to 224$\times$224 resolution.
In our experiments, we set $N_s$ and $N_f$ to 8 by default \tcb{unless otherwise specified}. The SF-Slots module generates $N= t^d \times N_s + M^d \times N_f = 8\times8 + 16\times8 = 192$ slots (tokens) from $16\times16\times T$ dense video tokens.

We pre-train the slot attention module of the Slow-Slots branch for 100 epochs and that of the Fast-Slots branch for 100 epochs, respectively. For the single branch instruction tuning, we tune the model powered by the Slow-Slots branch by 3 epochs and that powered by the Fast-Slots branch by 3 epochs. We jointly tune the two-branch model by 3 epochs. 
All models are trained using a single NVIDIA A100 GPU. 

The linear projection layer S-Proj., F-Proj. and Proj. consists of 1024, 1024, and 4096 neurons, respectively.
We adopt a network structure that is similar to the slot attention from \citet{singh2021illiterate} to build the slot attention modules, with learnable slot initialization.

More details please refer to Appendix \ref{subsec:training details}.

\subsection{Data Details and Evaluation Metrics}

We use the Video Instruction Data, collected by \citet{maaz2023video}, for video instruction tuning. This comprehensive dataset comprises approximately 100K video text (question-and-answer) pairs, which are generated from the ActivityNet dataset with an average video length of 180 seconds. The dataset is characterized by a diverse array of question types. 

We evaluate the performance on three open-ended video question-answering (QA) benchmarks like MSVD-QA\cite{chen2011collecting}, MSRVTT-QA\cite{xu2016msr}, and ActivityNet-QA \cite{caba2015activitynet}.

We evaluate models using accuracy (\%) and average score metrics by \citet{maaz2023video}, employing ChatGPT to judge prediction correctness. ChatGPT reviews each QA pair and issues a binary correctness verdict and a similarity score from 0 (least) to 5 (most) (See Appendix \ref{sec:metrics} for more details). 

\subsection{Comparison with the State-of-the-Arts}
We evaluate the performance of our Slot-VLM against the state-of-the-art methods on three zero-shot video QA benchmarks. Table~\ref{tab:SOTA} shows the results. All these models use Vicuna-7B \cite{vicuna2023} as the LLM. 

In the field of video-language modeling, there is no converged standard on the training datasets, including the data for pre-training and that for instruction tuning. For example, Video-LLaMA \cite{damonlpsg2023videollama} uses Webvid-2M short video dataset and CC595k image caption datasets for pre-training to enable video features to contain as much visual knowledge as possible; VideoChat2 \cite{li2023mvbench} performs per-training of two stages for vision-language alignment (15M image captions from CC3M and CC12M and 10M video captions form WebVid-10M) and vision-language connection (adding 2M image captions, and 10M video captions from InternVid \cite{wang2023internvid}), respectively. Following Video-ChatGPT \cite{maaz2023video}, we only perform instruction tuning. For instruction tuning, as listed in the second column of Table~\ref{tab:SOTA}, different methods use different data. 

We have the following observations/conclusions. \textbf{1)} In comparison to the methods highlighted in gray that utilize the same 100k video-text pairs from Video-ChatGPT for instruction tuning, our Slot-VLM consistently outperforms its competitors. It achieves superior accuracy, surpassing Video-ChatGPT \cite{maaz2023video} by \textbf{10\%} on MSVD-QA, \textbf{20.4\%} on MSRVTT-QA, and \textbf{13.1\%} on ActivityNet-QA. Furthermore, it exceeds Video-LLaVA \cite{lin2023video} by \textbf{10.1\%}, \textbf{11.4\%}, and \textbf{7.6\%}, and BT-Adapter \cite{liu2023one} by \textbf{7.9\%}, \textbf{18.5\%}, and \textbf{2.2\%} across the same benchmarks, respectively, demonstrating the high efficacy of our framework. \textbf{2)} An examination of two models from Video-LLaVA (rows five and six) reveals that augmenting the dataset with image-text pairs significantly benefits video reasoning. However, due to constraints in resources and for simplicity, we limit our instruction tuning to only the 100k video-text pairs. \textbf{3)} Slot-VLM secures either the best or second-best performance across all evaluated datasets, positioning it within the upper tier of models, even though the volume of vision-text pairs used for our training is considerably less than that of our counterparts. \tcb{\textbf{4)} Compared with Chat-UniVi \cite{jin2023chat} that leveraguses clustering to aggregate/compress tokens, our Slot-VLM outperforms it significantly by \textbf{9.9\%} on MSVD-QA, \textbf{15.1\%} on MSRVTT-QA, and \textbf{2.5\%} on ActivityNet-QA in accuracy. Note that Chat-UniVi uses 2 million video-text pairs and 433K image-text for instruction tuning while we use only 100K video-text pairs.}

\begin{table}[t]
\vspace{-1em}
  \centering
  \caption{Ablation studies on the effectiveness of our Slot-VLM. We compare our schemes powered by slot attention with the schemes powered by Q-Former under our framework.}
  \resizebox{0.49\textwidth}{!}{ 
    \begin{tabular}{c|c|cc|cc}
    \hline
    \multicolumn{2}{c|}{\multirow{2}[4]{*}{Model}} & \multicolumn{2}{c|}{In-domain} & \multicolumn{2}{c}{MSVD-QA} \bigstrut\\
\cline{3-6}    \multicolumn{2}{c|}{} & Acc.  & Score & Acc.  & Score \bigstrut\\
    \hline
    \multirow{2}[2]{*}{Slow branch } & Slow-Qformer-VLM & 38.8  & 2.45  & 70.9  & 3.67 \bigstrut[t]\\
          & Slow-Slot-VLM & \textbf{42.4} & \textbf{2.56} & \textbf{73.4} & \textbf{3.73} \bigstrut[b]\\
    \hline
    \multirow{2}[2]{*}{Fast branch} & Fast-Qformer-VLM & 38.4  & 2.43  & 72.2  & \textbf{3.70} \bigstrut[t]\\
          & Fast-Slot-VLM & \textbf{43.1} & \textbf{2.55} & \textbf{73.2} & 3.69 \bigstrut[b]\\
    \hline
    \multirow{2}[2]{*}{Two branches} & SlowFast-Qformer-VLM & 21.9  & 1.86  & 69.6  & 3.56 \bigstrut[t]\\
          & Slot-VLM & \textbf{48.8} & \textbf{2.75} & \textbf{74.9} & \textbf{3.76} \bigstrut[b]\\
    \hline
    \end{tabular}%
    }
  \label{tab:Ablation-QFormer}%
  \vspace{-1em}
\end{table}%

\subsection{Ablation Studies}
We study the effectiveness of our semantics-centric designs, SlowFast two branch design, and the hyper-parameter choices, on the test set of the Video Instruction Data from \citet{maaz2023video}, which we refer to as In-domain, and on MSVD-QA \cite{chen2011collecting}. 

We name our scheme with only the Slow-Slots branch as Slow-Slot-VLM, the scheme with only the Fast-Slots branch as Fast-Slot-VLM, and our final scheme as Slot-VLM.

\noindent\textbf{Effectiveness of using Semantics-centric Tokens} 
Under our framework, we replace our slot attention modules for generating reduced tokens by Q-Former \cite{li2023blip}, using similar training strategies. Similarly, we name the schemes with the slow branch, the fast branch, and two branches as Slow-QFormer-VLM, Fast-QFormer-VLM, and SlowFast-QFormer-VLM, respectively. 

We found that Q-Former
is not capable of
decoupling visual tokens to semantically meaningful instances (Please see Appendix~\ref{subsec:vis-QFormer} for the visualization of the attention maps). As shown in Table~\ref{tab:Ablation-QFormer}, we can see that \emph{the schemes using slot attention outperform those using Q-Former in both the single branch settings and the two branch setting.} Slow-Slot-VLM outperforms Slow-QFormer-VLM by 3.6\% on the indomain test data (In-domain) and 2.5\% on MSVD-QA, respectively. Fast-Slot-VLM outperforms Fast-QFormer-VLM by 4.7\% and 1.0\%. 

In addition, for the two-branch scheme SlowFast-QFormer-VLM, we found that tuning the Q-Formers and projection layers even with the single-branch trained parameters as initialization cannot lead to satisfactory results, which are poorer than the single-branch schemes. That may be caused by the difficulty in aligning the output tokens of the two branches. 

\noindent\textbf{Effectiveness of our Two-branch Design} Our scheme Slot-VLM with two branches outperforms our single-branch schemes Slow-Slot-VLM by 6.3\%/1.5\% on In-domain/MSVD-QA and Fast-Slot-VLM by 5.6\%/1.7\% in accuracy, respectively. Intuitively, our semantics-centric tokens from the two branches are easier to align with each other and align with LLM.

\noindent\textbf{Influence of Hyper-parameters} We study the influence of hyper-parameters and show the analysis in Appendix \ref{subsec:Hyperparameters}.

\subsection{Visualization}

\noindent\textbf{Object-Centric Slots}
We visualize the spatial cross-attention masks from our Slot-VLM for indicating the forming of object-centric slots for each frame, with two video samples shown in  Figure~\ref{fig:visualization-S-mark}. We can see that some meaningful slots have been formed. For example, in the first video, the slots marked by red, purple, green, and blue in the first video (left) correspond to ``background", ``human body", ``head", and ``barbell". 

For comparison, we also visualize the cross-attention masks of Q-Former from the scheme Slow-QFormer-VLM in Appendix \ref{subsec:Vis-Slow-QFormer}. We observe that our object-centric slots from Slot-VLM are better semantically decoupled.

Moreover, we found the instruction tuning can further promote the decoupling of slots (see Appendix~\ref{subsec:vis-stages}, where the well decoupled representations may better align with LLM. 

\noindent\textbf{Event-Centric Slots}
Figure~\ref{fig:Visualization-T-Sep}(a) visualizes the temporal cross-attention masks from our Fast-Slot branch for indicating the forming of event-centric slots for each spatial position (in total $h^d \times w^d = 16$ positions). Similarly, Figure~\ref{fig:Visualization-T-Sep}(b) visualizes the temporal cross-attention masks from Fast-QFormer-VLM for indicating the forming of query tokens for each spatial position. 

We can see that in our Slot-VLM, along the temporal axis, similar contents tend
to be allocated to the same slot. In other words, different slots capture different contents (events) and present decoupled semantics. In contrast, in the Fast-QFormer-VLM, different contents are usually assigned to the same query or are uniformly assigned to different queries. The queries learned from Q-Former do not present decoupled semantics.

\section{Conclusion}
In this work, we introduce a new framework, Slot-VLM, that aims to generate a small set of semantically decoupled video tokens to comply with LLM for effective video reasoning. Particularly, we design a dual-branch SlowFast Slots module to learn object-centric slots and event-centric slots to jointly capture the spatial object details and temporal dynamics. These semantic-centric slots are taken as the input to LLM for effective video reasoning. Experimental results demonstrate the superiority of our framework and show the effectiveness of using semantics-decoupled representations for aligning with LLM. However, our current representations are still far from perfect where object instances and events are not ideally segmented. 
We anticipate this work will inspire more investigations towards semantic-centric visual representations for video-language modeling.
\label{sec:conclusion}


\section*{Impact Statements}
This paper endeavors to make notable advancements in video-language modeling, a field integral to enhancing our interaction with and comprehension of video content. We aspire to drive positive innovation, yet we remain aware that technological progress may have unforeseen side effects. We call upon the users to use AI responsibly.


\bibliography{Slot-VLM}
\bibliographystyle{icml2024}

\newpage
\appendix
\onecolumn

\section{More Details about Evaluation Metrics}
\label{sec:metrics}
In our study, we adopt the evaluation metrics of accuracy and average score as established by \citet{maaz2023video}. To assess the accuracy of our model's predictions, we utilize ChatGPT (chatgpt35-turbo) as an evaluator. ChatGPT processes each question alongside the corresponding ground truth and the model's predicted answer. It then provides a binary "yes" or "no" judgment on the correctness of the prediction for accuracy assessment. Additionally, ChatGPT assigns an integer score ranging from 0 to 5 to quantify the closeness of the predicted answer to the ground truth, with 0 indicating no similarity and 5 denoting a close match.

\section{More Implementation Details about Three Stages Training}
\label{subsec:training details}

\textbf{Stage-1} We reconstruct the features (rather than images) obtained by CLIP ViT-L/14 \tcb{using a transformer decoder as in} DINOSAUR\cite{seitzer2022bridging}. We train the models for 100 epochs with a learning rate 1e-4. We use the Slot Attention module in SLATE\cite{singh2021illiterate}. \tcb{The dimension of each slot is set to 1024.} 

\textbf{Stage-2} We use the Video Instruction Data proposed by \citet{maaz2023video} to train our single branch schemes, respectively.
The Fast-Slot branch/Slow-Slot branch and the projection layer Proj. are tuned. We set the number of epochs to 3, the learning rate to 2e-5. \tcb{We adopt the cosine annealing learning rate}. We set the batch size to 40 and train on a single A100 GPU.

\textbf{Stage-3} We load the parameters of SF-Slots module obtained from the second stage, and use the Video Instruction Data for instruction tuning. The Fast-Slot branch/Slow-Slot branch and the projection layer Proj. are tuned. The training hyper-parameters used in this stage are the same as those in the second stage.

\section{More Ablation Studies}
\label{subsec:Hyperparameters}
\noindent\textbf{Influence of High Spatial Resolution for the Slow-Slots Branch and High Frame Rate for the Fast-Slots Branch} We study the influence of spatial resolution of the visual features in the Slow-Slots branch and the influence of frame rate of the visual features in the Fast-Slots branch, respectively. As shown in Table~\ref{tab:Ablation-Spatial-Resolution}, when we reduce the spatial resolution from $16 \times 16$ to $4\times4$, the performance drops by 1.4\% on MSVD-QA in accuracy. When we reduce the frame rate by a factor of 8 from 1fps to 1/8fps, the performance drops by 1.2\% on MSVD-QA in accuracy. The difference on the in-domain test is small. That may be because most of the questions in the test set are high level questions which ask to summarize the video, which do not require fine-granularity details. 

\begin{table}[h]
  \centering
  \caption{Ablation study on the influence of high spatial resolution for the Slow-Slots branch and high frame rate for the Fast-Slots branch. Slow-Slot-VLM ($4\times4$) denotes the spatial resolution is reduced from $16\times16$ to $4\times4$. Fast-Slot-VLM ($T/8$) denotes the frame rate is reduced by a factor of 8.}
    \resizebox{0.49\textwidth}{!}{
    \begin{tabular}{c|c|cc|cc}
    \hline
    \multicolumn{2}{c|}{\multirow{2}[4]{*}{Model}} & \multicolumn{2}{c|}{In-domain} & \multicolumn{2}{c}{MSVD-QA} \bigstrut\\
\cline{3-6}    \multicolumn{2}{c|}{} & Acc.  & Score & Acc.  & Score \bigstrut\\
    \hline
    \multirow{2}[2]{*}{Slow branch } & Slow-Slot-VLM ($4\times4$) & 42.1  & 2.52  & 72.0  & 3.64 \bigstrut[t]\\
          & Slow-Slot-VLM & \textbf{42.4} & \textbf{2.56} & \textbf{73.4} & \textbf{3.73} \bigstrut[b]\\
    \hline
    \multirow{2}[4]{*}{Fast branch} & Fast-Slot-VLM ($T/8$) & 42.9  & 2.54  & 72.0  & 3.64 \bigstrut[t]\\
         & Fast-Slot-VLM & \textbf{43.1} & \textbf{2.55} & \textbf{73.2} & \textbf{3.69} \bigstrut[b]\\
    \hline
    \end{tabular}%
    }
  \label{tab:Ablation-Spatial-Resolution}%
\end{table}%

\noindent\textbf{Influence of Hyperparameter $N_s$} \tcb{We study the influence of the number of object-centric slots $N_s$ of a frame and show the results in Table~\ref{tab:abla-slot-num}. We can see that as the increase of the number of slots from 4 to 16 for a frame, the performance increases. }

\noindent\textbf{Influence of Hyperparameter $N_f$} \tcb{We study the influence of the number of event-centric slots $N_f$ of a spatial position and show the results in Table~\ref{tab:abla-slot-num}. We can see that as the increase of the number of slots from 4 to 16 for spatial position, the performance increases. }

The total number of tokens when $N_s=8$ and $N_f=8$ is 192. In comparison, the total number of tokens when $N_s=16$ and $N_f=16$ is 768, which increases the inference complexity. To trade off the performance and complexity, we use $N_s=8$ and $N_f=8$ by default.

\begin{table}[htbp]
  \centering
  \caption{Ablation study on the influence of the number of object-centric slots $N_s$ and the influence of the number of event-centric slots $N_f$, respectively.}
  \resizebox{0.49\textwidth}{!}{
    \begin{tabular}{c|c|cc|cc}
    \hline
    \multicolumn{2}{c|}{\multirow{2}[4]{*}{Model}} & \multicolumn{2}{c|}{In-domain} & \multicolumn{2}{c}{MSVD-QA} \bigstrut\\
\cline{3-6}    \multicolumn{2}{c|}{} & Acc.  & Score & Acc.  & Score \bigstrut\\
    \hline
    \multirow{3}[2]{*}{Slow branch } & Slow-Slot-VLM (4slots) & 41.0  & 2.5   & 73.6  & 3.68 \bigstrut[t]\\
          & Slow-Slot-VLM (8slots) & 42.4  & 2.56  & 73.4  & 3.73 \\
          & Slow-Slot-VLM (16slots) & \textbf{45.4} & \textbf{2.64} & \textbf{74.7} & \textbf{3.73} \bigstrut[b]\\
    \hline
    \multirow{3}[2]{*}{Fast branch} & Fast-Slot-VLM (4slots) & 42.9  & 2.55  & 74.2  & 3.69 \bigstrut[t]\\
          & Fast-Slot-VLM (8slots) & 43.2  & 2.55  & 73.2  & 3.69 \\
          & Fast-Slot-VLM (16slots) & \textbf{46.0} & \textbf{2.59} & \textbf{74.9} & \textbf{3.69} \bigstrut[b]\\
    \hline
    \end{tabular}%
    }
  \label{tab:abla-slot-num}%
\end{table}%

\section{More Visualization}

\begin{figure*}[t]  
  \centering  
  \begin{subfigure}{1\linewidth}  
    \includegraphics[width=\linewidth]{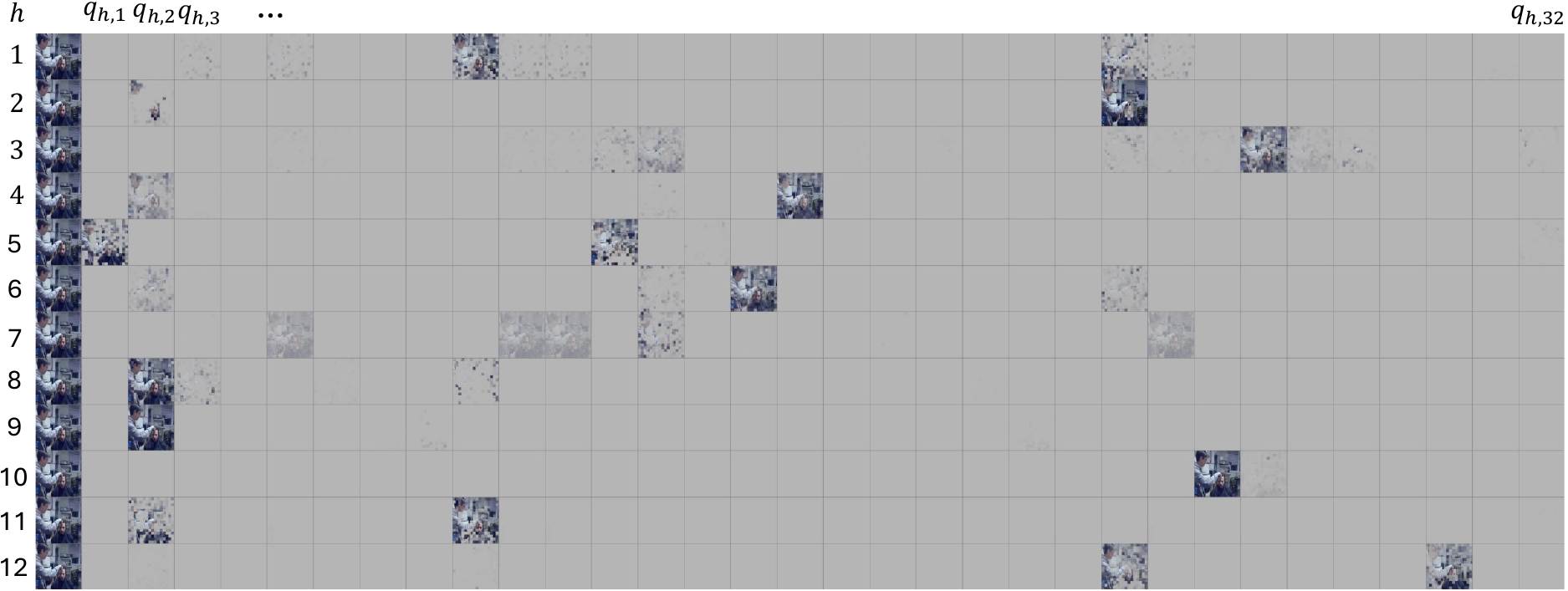}  
    \caption{ }  
    \label{fig:sub1}  
  \end{subfigure}  
  \hfill 
  \begin{subfigure}{1\linewidth}  
    \includegraphics[width=\linewidth]{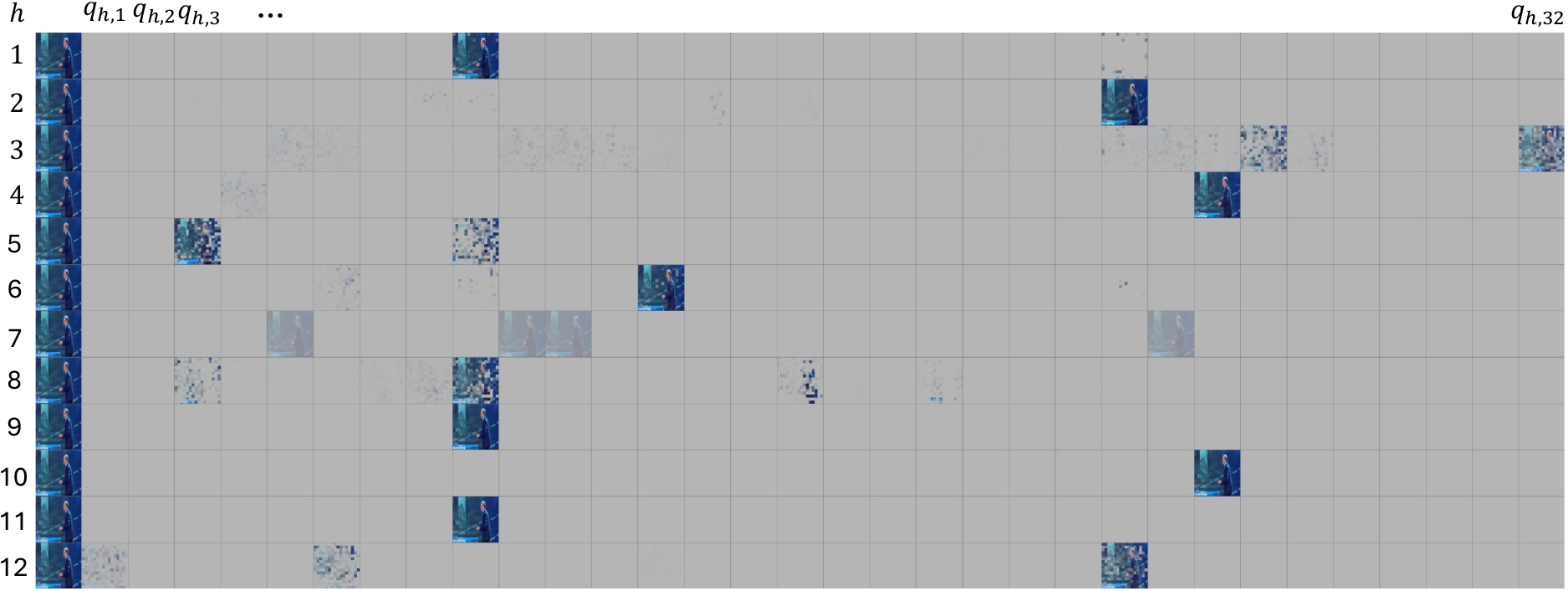}  
    \caption{ }  
    \label{fig:sub2}  
  \end{subfigure}  
  \caption{Visualization of spatial attention masks from the Q-Former in BLIP2 for two images in (a) and (b) respectively. We show the learned query masks for the 12 heads in 12 rows, respectively. In each row, we show the masks for the 32 queries. Note that the first column show the original image repeated by 12 times. There is no obvious evidence that different queries have learned decoupled semantics. }  
  \label{fig:Visualization-BLIP2}  
\end{figure*}  

\subsection{Visualization of Q-Former Attention Maps from BLIP2}
\label{subsec:vis-QFormer}
BLIP2 \cite{li2023blip} is a representative visual language pre-trained model that aggregates image features (from CLIP encoder) by a Q-Former into 32 learnable query tokens.
Q-Former employs a 12-head cross-attention mechanism to aggregate visual features \tcb{into a set of queries}, which \tcb{subsequently processes the queries} by a self-attention module and a feed-forward layer.
We analyze whether such well-trained Q-Former has learned decoupled semantics for the 32 query tokens. Figure~\ref{fig:Visualization-BLIP2} visualizes the learned cross-attention masks (laid out on the original image) corresponding to the 32 queries (32 columns, $\mathbf{q}_{h,1}, \mathbf{q}_{h,2}, \ldots, \mathbf{q}_{h,32}$) from 12 heads (shown in 12 rows, $h=1, 2, \ldots, 12$) for two images in (a) and (b) respectively. We can see that a query focuses on different regions for different heads. For a head, usually the information was allocated into only a few queries. However, there is no obvious evidence that different queries have learned decoupled semantics. In contrast, as shown in Figure~\ref{fig:visualization-S-mark}, our learned spatial slots have more remarkable decoupled semantics.

\subsection{Visualization of Q-Former Spatial Attention Maps from Slow-QFormer-VLM}
\label{subsec:Vis-Slow-QFormer}
\begin{figure*}[th]
  \centering
   \includegraphics[width=1\linewidth]{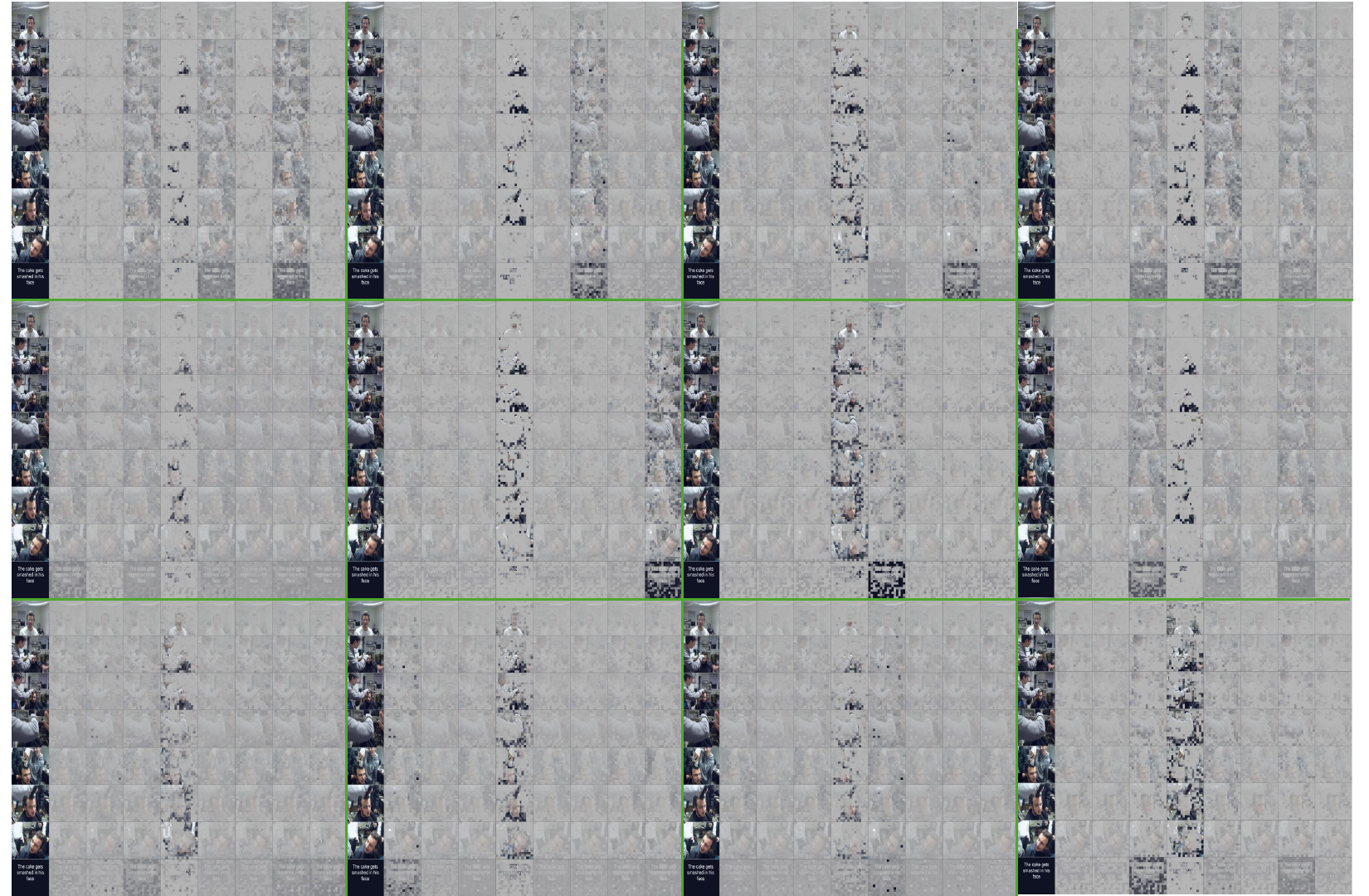}
   \caption{Visualization of spatial attention masks from our baseline scheme Slow-QFormer-VLM for 8 frames of a video. Since there are 12 heads in Q-Former, we plot 12 (3 $\times$ 4) sets of attention masks separated by green lines. For each head, we have $t^d=8$ frames as shown in each row; the first column shows the original frame; the second to the ninth columns show the cross attention mask (from Q-Former) for the 8 queries. We can see that the queries are not clearly decoupled, with each one being a mixture of spatial features. A feature in a spatial position is usually allocated to a couple of queries. Compared with the spatial attention masks from our Slot-VLM as shown in Figure~\ref{fig:visualization-S-mark}, it is less semantically decoupled for the learned queries. } 
   \label{fig:Visualization-SlowQFormer}
\end{figure*}

We visualize the spatial attention maps of the learned queries of Q-Former from our Slow-QFormer-VLM. Q-Former uses cross-attention of 12 heads to aggregate the visual features. In Figure~\ref{fig:Visualization-SlowQFormer}, we show the masks for 12 heads separately as shown in the $3 \times 4$ grids separated by green lines. For each head, we show all the 8 frames in different rows; the first column shows the original frame; the second to the ninth columns show the cross attention mask (from Q-Former) for the 8 queries. We can see that the queries are not clearly decoupled, with each one being a mixture of spatial features. A feature in a spatial position is usually allocated to a couple of queries instead of one. Compared with the spatial attention masks from our Slot-VLM as shown in Figure~\ref{fig:visualization-S-mark}, it is less semantically decoupled for the learned queries. 

\subsection{A Glimpse of the Original Video Used in Figure~\ref{fig:Visualization-T-Sep} and Enlarged Visualization}
\label{subsec:video-glimpse-temp} 
  
\begin{figure*}[th]
  \centering
   \includegraphics[width=1\linewidth]{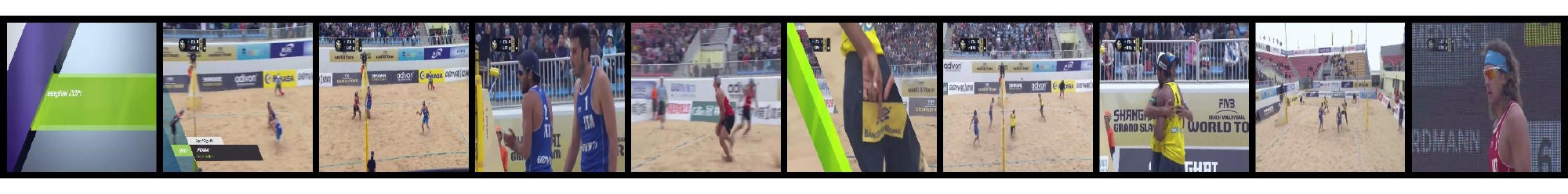}
   \caption{A glimpse of the original video used in Figure~\ref{fig:Visualization-T-Sep}. For visualization purpose, we only show the frames down-sampled at a factor of 8, which is 1/8fps.} 
   \label{fig:Temporal-OriginalFrames}
\end{figure*}

Figure~\ref{fig:Temporal-OriginalFrames} shows a glimpse of the original video used in Figure~\ref{fig:Visualization-T-Sep}.

\begin{figure*}[t]
  \centering
   \includegraphics[width=0.35\linewidth]{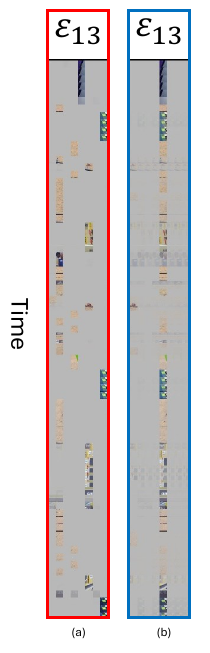}
   \caption{Enlarged visualization of temporal attention mask for the $13^{th}$ spatial position from (a) our Fast-Slots branch and (b) Fast-QFormer-VLM, respectively. \tcb{For simplicity, we also refer to slot as query here.} For the $13^{th}$ spatial position, we denote the set of learned temporal queries by $\mathcal{E}_{13}$. For this spatial position, the models generate $N_f=8$ queries by aggregating the temporal visual tokens.
   \tcb{The attention masks for $\mathcal{E}_{13}$ are denoted by a map of $T$ rows and $N_f=8$ columns,}
   with the visibility indicating which queries this temporal
   position belongs to.
   The higher the visibility, the greater the affinity between this temporal position and the query.
   We can see that in our Slot-VLM, similar contents tend
   to be allocated to the same slot, \ieno, different slots capture different contents (events) and present decoupled semantics. In contrast, in Fast-QFormer-VLM, different contents are usually assigned to the same query or are uniformly assigned to different queries.
   } 
   \label{fig:Visualization-T-Sep-Enlarged}
\end{figure*}

Figure~\ref{fig:Visualization-T-Sep-Enlarged} shows the enlarged visualization of attention masks for the $13^{th}$ spatial position that is previously shown in Figure~\ref{fig:Visualization-T-Sep}.   

\subsection{Visualization of Spatial Attention Maps for Spatial Slots from Stage-1 and Stage-2}
\label{subsec:vis-stages}

\begin{figure*}[t]
  \centering
   \includegraphics[width=1\linewidth]{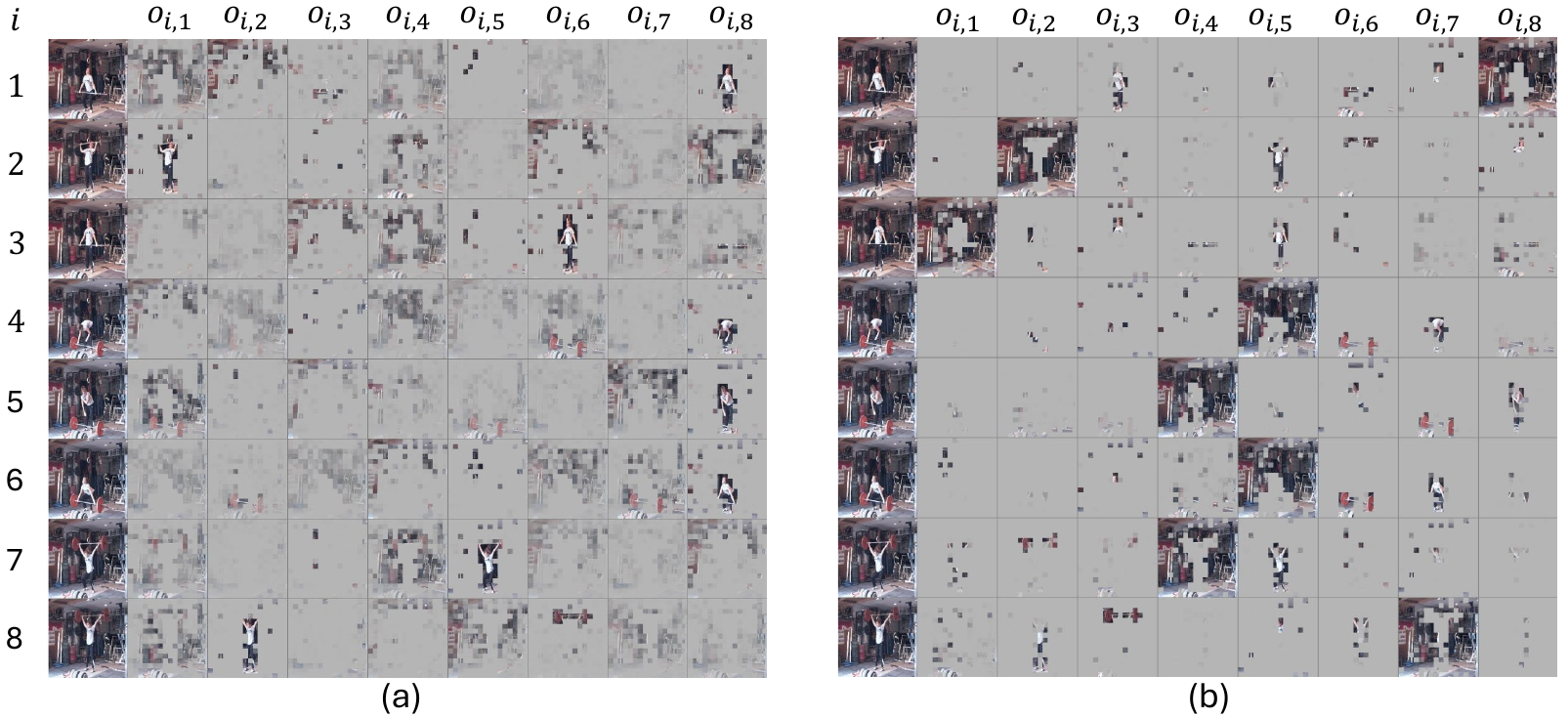}
   \caption{Visualization of spatial attention masks from (a) the stage 1 pre-training, and (b) the stage 2 after instruction tuning. We have $t^d=8$ frames as shown in 8 rows, indexed by $i$, where $i=1,\ldots,t^d$, respectively. The first column shows the original frame. The second to the ninth columns show the cross attention mask (from slot attention) for the $N_s=8$ object-centric slots $\mathcal{O}_i = \{\mathbf{o}_{i,1}, \ldots, \mathbf{o}_{i,N_s}\}$. Interestingly, we can see that after the instruction tuning, the learned slots are much more decoupled, where a spatial position usually contributes to multiple slots in stage 1 but only contributes to a very few slots in stage 2. } 
   \label{fig:Visualization-S-Stages-1}
\end{figure*}

\begin{figure*}[t]
  \centering
   \includegraphics[width=1\linewidth]{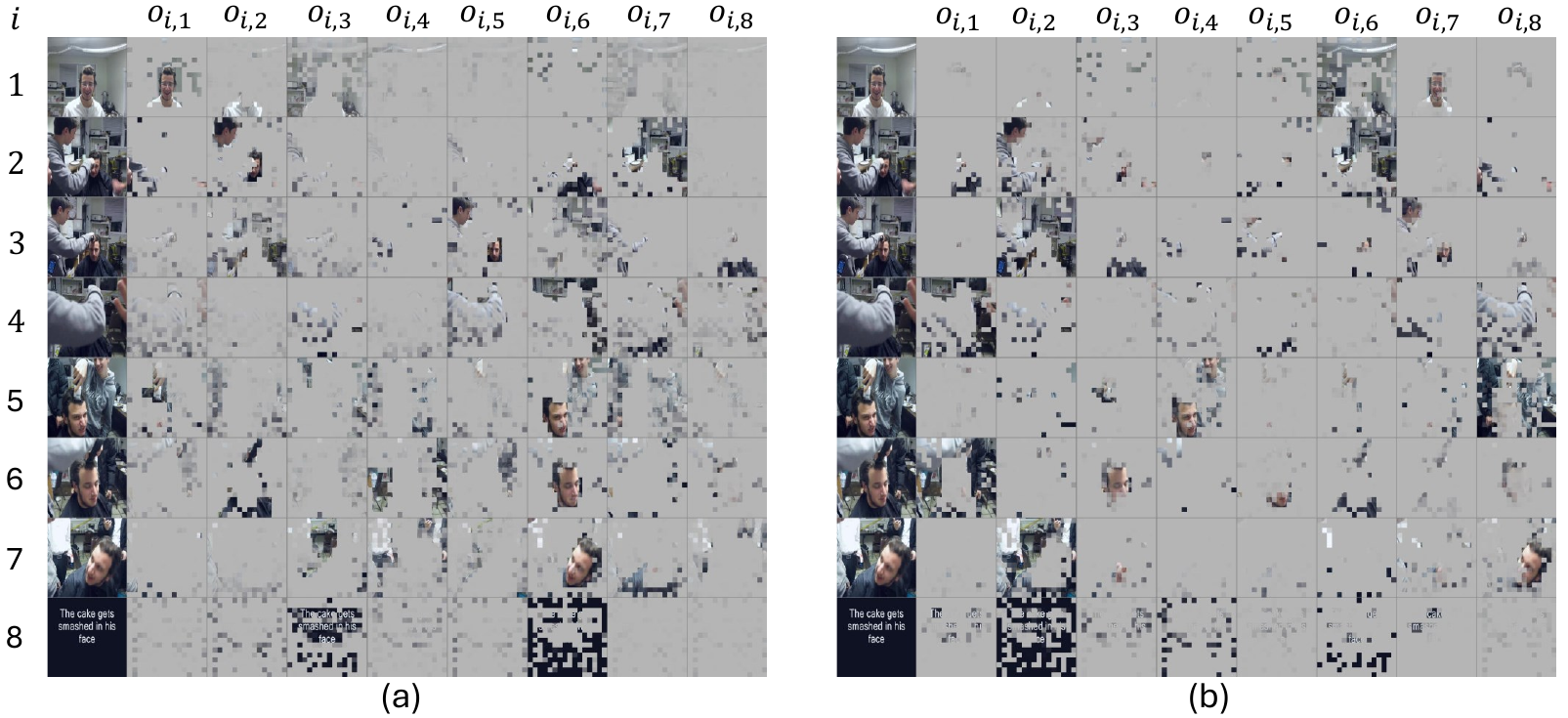}
   \caption{Visualization of spatial attention masks from (a) the stage 1 pre-training, and (b) the stage 2 after instruction tuning. We have $t^d=8$ frames as shown in 8 rows, indexed by $i$, where $i=1,\ldots,t^d$, respectively. The first column shows the original frame. The second to the ninth columns show the cross attention mask (from slot attention) for the $N_s=8$ object-centric slots $\mathcal{O}_i = \{\mathbf{o}_{i,1}, \ldots, \mathbf{o}_{i,N_s}\}$. Interestingly, we can see that after the instruction tuning, the learned slots are much more decoupled, where a spatial position usually contributes to multiple slots in stage 1 but only contributes to a very few slots in stage 2. } 
   \label{fig:Visualization-S-Stages-2}
\end{figure*}

In the first stage pre-training, slot attention is learned by reconstructing the features. In the second stage, LLM is incorporated for video instruction tuning. Here, we take spatial slots as examples to compare the slot attention masks from stage 1 and stage 2. Figure~\ref{fig:Visualization-S-Stages-1} and Figure~\ref{fig:Visualization-S-Stages-2} show the visualization for two examples, respectively. Interestingly, we observe that after the instruction tuning, the learned slots are better decoupled, where a spatial position usually contributes to multiple slots in stage 1 but only contributes to a very few slots in stage 2. Similar phenomenons are obtained for temporal slots. 

\end{document}